\pdfoutput=1

\documentclass[11pt]{article}

\usepackage{EMNLP2023}

\usepackage{times}
\usepackage{latexsym}
\usepackage{amsmath}
\usepackage[T1]{fontenc}

\usepackage[utf8]{inputenc}
\usepackage[]{mdframed}
\usepackage{framed}
\usepackage{microtype}

\usepackage{inconsolata}

\usepackage{mathtools}
\usepackage{booktabs}
\usepackage{enumitem}
\usepackage{hhline}
\usepackage{multirow}

\newcommand{\dataset}{\textsc{MemeCap}}

\definecolor{Green}{rgb}{0.05, 0.5, 0.06}
\definecolor{Purple}{rgb}{0.56, 0.0, 1.0}
\definecolor{Orange}{rgb}{1.0, 0.55, 0.0}\definecolor{Blue}{rgb}{0.0, 0.2, 0.4}

\title{\dataset{}: A Dataset for Captioning and Interpreting Memes}

\author{EunJeong Hwang$^{1,2}$ and Vered Shwartz$^{1,2}$ \\
$^1$ University of British Columbia~~~$^2$ Vector Institute for AI\\
{\tt \{ejhwang,vshwartz\}@cs.ubc.ca}}

\begin{document}
\maketitle
\begin{abstract}
Memes are a widely popular tool for web users to express their thoughts using visual metaphors. Understanding memes requires recognizing and interpreting visual metaphors with respect to the text inside or around the meme, often while employing background knowledge and reasoning abilities. We present the task of meme captioning and release a new dataset, \dataset{}. Our dataset contains 6.3K memes along with the title of the post containing the meme, the meme captions, the literal image caption, and the visual metaphors. Despite the recent success of vision and language (VL) models on tasks such as image captioning and visual question answering, our extensive experiments using state-of-the-art VL models show that they still struggle with visual metaphors, and perform substantially worse than humans.  
\end{abstract}

\section{Introduction}
\label{sec:intro}
\begin{figure}[!t]
    \centering\includegraphics[width=0.45\textwidth]{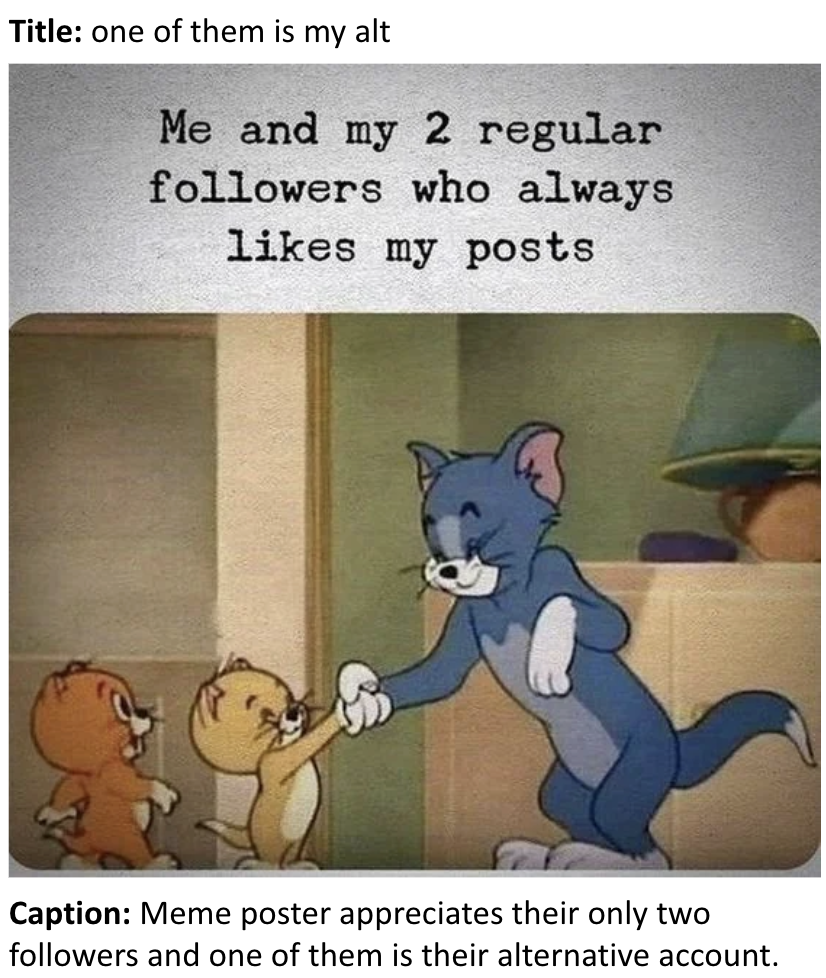}
    \caption{A meme and its title. The caption describes what the meme poster was trying to convey.}
    \label{fig:meme_ex}
\end{figure}

Web users frequently communicate their thoughts and feelings online using memes \cite{buchel2012internet, tanaka-etal-2022-learning}. Memes are created by taking an existing widespread image and attaching new meaning to it by altering the text inside the image. For example, in Figure~\ref{fig:meme_ex}, Tom cat is a metaphor for the person who posted the meme and the cats he is shaking hands with represent his two regular followers who always like his posts. This incongruity between the image and the text makes memes humorous \cite{tanaka-etal-2022-learning}. 

Because of their complementary nature, interpreting the meaning of a meme requires understanding both the visual and text modalities. Moreover, memes are often posted on social media platforms along with additional text, such as ``one of them is my alt'' in Fig.~\ref{fig:meme_ex}, which is further needed to understand the meme.

Recently, there is a surge of vision and language (VL) models \cite[e.g.][]{flamingo,blip2,gpt4}. VL models have shown remarkable capabilities in generating detailed and accurate descriptions of images in both zero-shot and in-context setups. Such models are first pre-trained on language-only and vision-only datasets, and then trained on tasks such as image captioning and visual question answering, where the redundancy between the vision and language is used to embed them in a shared space. For example, the majority of image captions in existing datasets describe what is depicted in the image, at most adding subjective interpretations or inferences about the story behind the image  \cite{alikhani-etal-2020-cross}. In contrast, there is little work on visual metaphors to date \cite{zhang-etal-2021-multimet,visual-metaphors}.

In this paper, we are investigating whether VL models can successfully interpret memes. We propose the task of \emph{meme captioning}, in which models are presented with a meme along with its title (e.g. the title of the post containing the meme), and is tasked with generating a concise caption describing the meaning of the meme. This task goes beyond object recognition and language understanding. It is challenging due to the metaphorical role of the visual content of the meme \cite{scott2021memes}. For example, in Fig.~\ref{fig:meme_ex}, the model needs to recognize that Tom cat is merely a metaphor for the meme poster, and that handshaking signals appreciation. The literal content of the image, such as Tom or the handshake, should not be part of the meme caption. Recognizing and interpreting such metaphors involve detecting facial expressions, the tone expressed in the texts, making commonsense inferences, and more  \cite{bittonguetta2023breaking}. 

To that end, we collected a meme captioning dataset \dataset{}, containing 6,384 memes along with their captions. Each meme is also annotated with the literal image description (e.g. ``Tom cat is shaking hands with two small cats and smiling''), and the visual metaphors (e.g. Tom is a metaphor for the meme poster). 

We establish comprehensive baseline performances with recent large-scale VL models, in various training setups (e.g. zero-shot, few-shot, fine-tuning), and inputs (i.e. meme, title, literal image captions, and metaphors). Human evaluation of the generated captions shows that models are far from humans in captioning memes. In particular, models tend to ignore important visual or textual elements, and instead, repeat the text inside the meme or make up fake elements. Our findings merit future research on this task.\footnote{Our code and data are available at:\\ \url{https://github.com/eujhwang/meme-cap}}

\section{Background}
\label{sec:bg}
\subsection{Metaphors}
\label{sec:bg:metaphors}

Most work on metaphors is limited to textual metaphors, and pertains to collecting resources \cite{dodge-etal-2015-metanet}, detecting or interpreting metaphorical expressions in context \cite{choi-etal-2021-melbert,chakrabarty-etal-2021-figurative,aghazadeh-etal-2022-metaphors, chakrabarty-etal-2022-rocket}, and metaphor generation \cite{stowe-etal-2021-exploring,chakrabarty-etal-2021-mermaid}. 

Recently, there has been interest in visual metaphors. Visual metaphors occur when a target concept is compared to another visual element (vehicle) \cite{forceville1996pictorial}. MultiMET \cite{zhang-etal-2021-multimet} and Met-Meme \cite{xu2022met} are two datasets of text-image pairs with annotations for the existence and types of metaphors, sentiment, and more. \newcite{visual-metaphors} tested image generation models on prompts involving a visual metaphor such as ``My bedroom is a pigsty''. They found the unsatisfactory performance can be improved by using a large language model (LLM) to interpret the visual metaphors and add details to the prompt, such as ``messy bedroom''.

\subsection{Memes}
\label{sec:bg:memes}

Prior work on memes focused on detecting hateful or harmful content in memes \cite{qu2022disinfomeme, sharma-etal-2023-characterizing, hateful-memes}, classifying memes to humorous or not \cite{tanaka-etal-2022-learning}, analyzing the sentiment of memes \cite{chhavi2020memotion}, and generating memes (e.g. the ImgFlip575K dataset).\footnote{\url{https://github.com/schesa/ImgFlip575K_Dataset}} 

Although MultiMET \cite{zhang-etal-2021-multimet} does not focus specifically on memes, the images were collected from a range of sources including
social media, which contains memes. The similar Met-Meme dataset \cite{xu2022met} focuses on memes. Differently from our work, both datasets contain annotations for visual metaphors while \dataset{} also contains meme captions. 

\subsection{Other Image Datasets}
\label{sec:bg:other}

The WHOOPS benchmark \cite{bittonguetta2023breaking} consists of unconventional human-created and machine-generated images that defy commonsense (e.g. an image of ``Albert Einstein holding a smartphone''), along with their textual descriptions. It's meant to be used for image captioning, image-text matching, visual question answering, and explanation generation. In contrast, our work focuses on memes, and tests models on their ability to interpret real memes posted by web users. 

\section{The \dataset{} Dataset}
\label{sec:dataset}
\begin{figure*}[!t]
\centering\includegraphics[width=\textwidth]{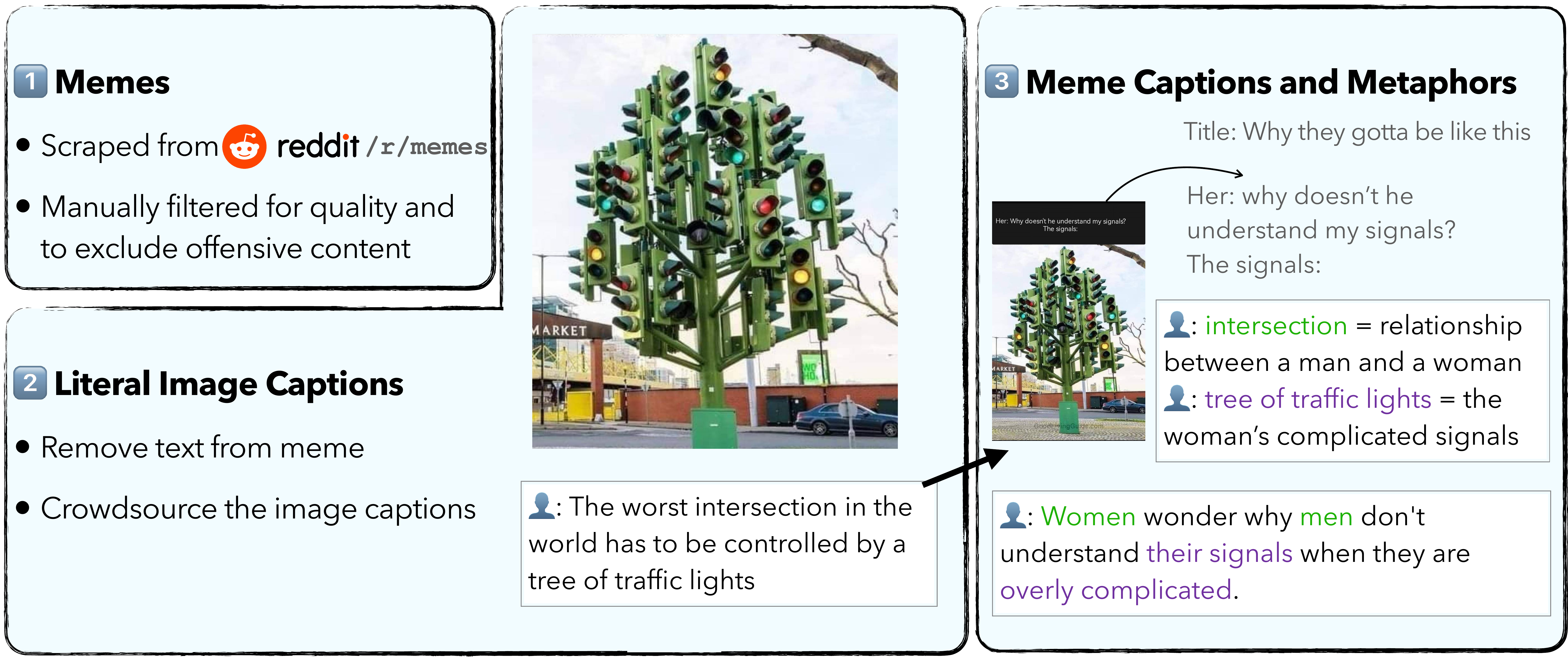}
    \caption{Overall process of collecting memes, literal image captions, visual metaphors, and meme captions.}
    \label{fig:anno-process}
\end{figure*}
\begin{table}[!t]
    \small
    \centering
    \begin{tabular}{lrrrr}
    \toprule
   & \#Memes & \#M-Cap & \#I-Cap & \#Mph \\ \midrule
  Train+Val & 5,828  & 1.0 & 1.0 & 2.1 \\ 
  Test & 559  & 3.4 & 1.0 & 3.1 \\ \bottomrule
    \end{tabular}
    \caption{The number of memes in \dataset{}, and the average number of meme captions (M-Cap.), image captions (I-Cap.), and  metaphorical keywords (Mph) per meme.}
    \label{tab:dataset-stat}
\end{table}

 The overall data collection and annotation process is illustrated in Figure~\ref{fig:anno-process}. We collected memes (Sec~\ref{sec:dataset:memes}) and crowdsourced their captions (Sec~\ref{sec:dataset:captions}). We present the data splits and statistics in Sec~\ref{sec:dataset:stats}. 

\subsection{Memes}
\label{sec:dataset:memes}

We scraped memes from Reddit using the publicly available API.\footnote{\url{https://www.reddit.com/dev/api/}} In particular, we focused on the subreddit \texttt{/r/memes} and collected posts that contained a meme with a post title. To ensure that the text and image are complementary, we manually examined the memes and excluded memes that lacked any text or contained an excessive number of characters. To exclude offensive content from the dataset, we filtered out memes with profanity in the text using the Google banned word list.\footnote{\scriptsize\url{https://github.com/coffee-and-fun/google-profanity-words}} We also filtered out images with sexual content, for which the NudeNet Classifier returned an unsafe score higher than 0.9.\footnote{\url{https://github.com/notAI-tech/NudeNet}} 

\begin{figure*}[!t]
    \centering\includegraphics[width=\textwidth]{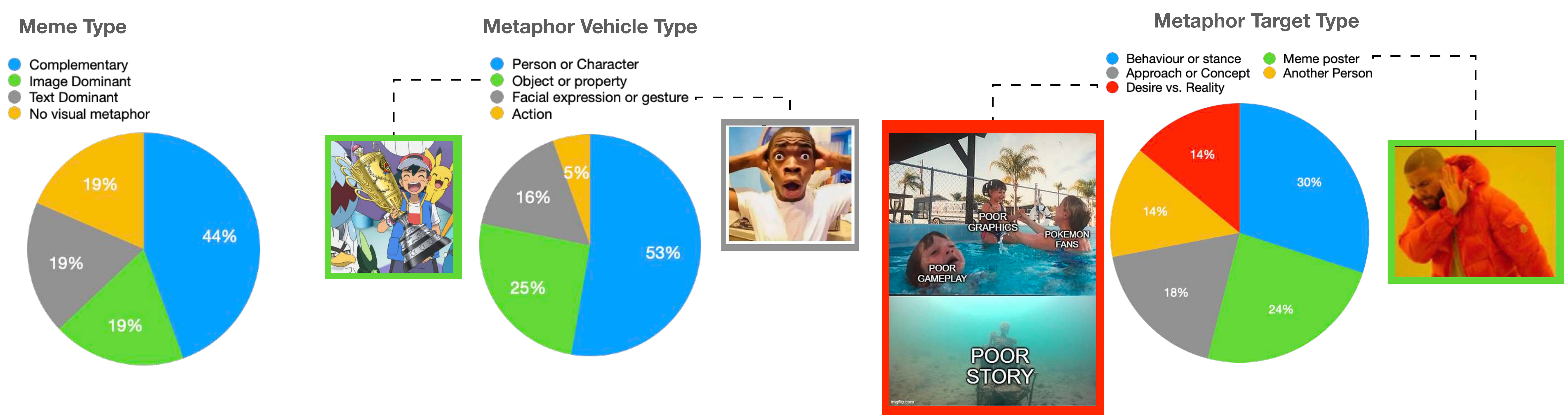}
    \caption{(1) \textbf{Meme Type}: Percent of memes with no visual metaphors, and with metaphors that can be understood with the text alone, vision alone, or both (complementary). (2) \textbf{Metaphor Vehicle Type}: Types of visual elements used to convey a metaphorical meaning. (3) \textbf{Metaphor Target Type}: The intended meanings of the metaphors.}
    \label{fig:metaphor_analysis}
\end{figure*}
    
\subsection{Captions}
\label{sec:dataset:captions}

We conducted two rounds of annotations to obtain the captions. In the first round, we collected the literal image descriptions, disregarding the text in the memes, while in the second round, we collected the meme caption along with the visual metaphors.

\paragraph{Literal Image Captions.} We asked workers to caption the image, disregarding the text. For example, a suitable literal image caption for Figure~\ref{fig:meme_ex} is ``Tom cat is shaking hands with two small cats and smiling''. To prevent biasing the workers with the text inside the meme, we identified and removed the text in the meme using the LaMa inpainting tool  \cite{suvorov2021resolution}. We collected one caption for each meme, which we manually verified. 

\paragraph{Meme Captions.} We showed a second set of annotators the full meme, title, and the literal image caption, and asked them to provide a meme caption. This HIT included two steps. First, workers were asked to indicate for each term in the literal image caption whether it was used metaphorically, and if so, what was the target of the metaphor (e.g., ``Tom cat'' is a metaphor for the meme poster). We then instructed the workers to write a concise caption describing the meaning that the meme poster was trying to convey, while excluding the metaphor vehicles (e.g., not mentioning Tom). We collected one caption for each meme in the training set, and 2 to 4 captions for memes in the test set.

\paragraph{}
Both rounds of annotations were conducted on Amazon Mechanical Turk (MTurk). To ensure the quality of annotations, we required that workers were located in English speaking countries (US, UK, Canada, Australia, and New Zealand), had an acceptance rate of at least 98\% on 5,000 prior HITs, and passed a qualification test similar to the task. We paid \$0.03 for the image captioning task and \$0.16 for the meme captioning task.

We excluded from the dataset any memes that workers in each of the rounds marked as offensive, sexual, hateful, or uninterpretable. 

\subsection{Final Dataset}
\label{sec:dataset:stats}

We clustered the examples in the dataset based on the vector representation of their meme captions using OPT2.7b \cite{opt}. To ensure the diversity of topics in both the training and test sets, we then sampled 10\% of the memes from each cluster and assigned them to the test set, and the rest of the memes into the training and validation set.\footnote{Note that our dataset doesn't contain duplicate memes.} Table~\ref{tab:dataset-stat} shows the statistics of our dataset.

\subsection{Types of Metaphors}
\label{sec:analysis:metaphor_types}

We manually analyzed 28 memes along with their metaphor annotations. 

\paragraph{Meme Type.} First, following \newcite{zhang-etal-2021-multimet} and \newcite{xu2022met}, we categorized the memes into three categories: \emph{text dominant} and \emph{image dominant}, where the text or the image respectively may be enough to understand the metaphor, and \emph{complementary}, where both modalities are required. We added a fourth category for memes that had no metaphor, i.e. whose meaning is conveyed explicitly in the text. The left part of Figure~\ref{fig:metaphor_analysis} shows that the 44\% of memes are complementary, but each of the other categories is also prominent with 19\%. 

\paragraph{} We then looked at the human annotations we obtained in Sec~\ref{sec:dataset:captions} for the metaphors in each meme. We looked at the vehicle, i.e. the visual element used to convey the metaphorical meaning, as well as the target, i.e. the meaning itself. 

\paragraph{Metaphor Vehicle Type.} The middle part of Fig~\ref{fig:metaphor_analysis} shows that the most common vehicle is a person or a character, followed by objects (such as the trophy), facial expressions or gestures (such as the surprised look on the man's face), and actions. 

\paragraph{Metaphor Target Type.} The types of targets are displayed in the right part of Fig~\ref{fig:metaphor_analysis}. The majority of the metaphors describe either a behavior or stance towards a certain topic, or the meme poster themselves (with a person vehicle, such as Drake). Other categories are an approach or a concept (for which the meme poster expresses a certain stance), another person, and a ``desire vs. reality'' meme such as the drowning meme illustrated in Fig~\ref{fig:metaphor_analysis}. 

\section{Experimental Setup}
\label{sec:exp_setup}
\begin{figure*}[!t]
\centering\includegraphics[width=\textwidth]{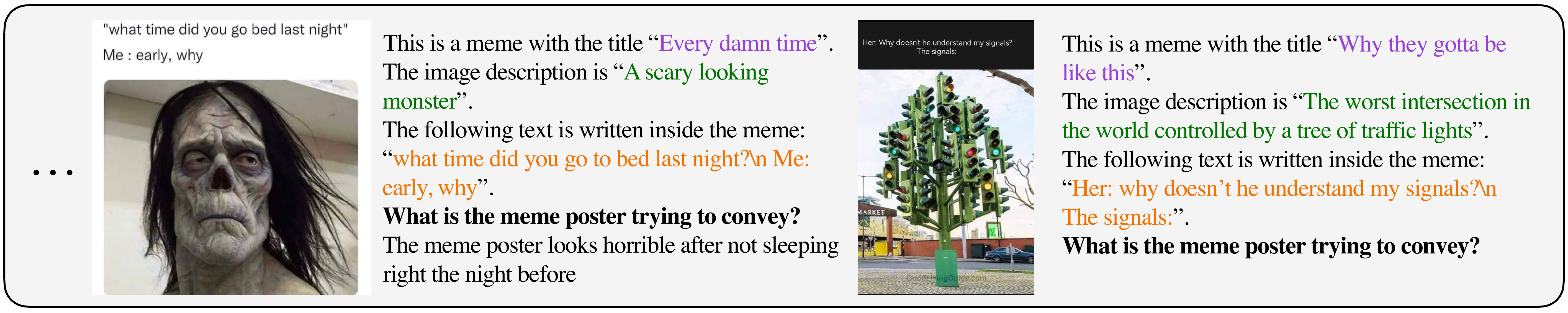}
    \caption{An example of the few-shot setup with the following inputs: meme, image description, and the text inside the meme. The figure shows the last in-context meme and the target meme.}
    \label{fig:few_shot}
\end{figure*}

We report the performance of various baselines on \dataset{}. All models are tasked with generating a meme caption, and are based on pre-trained VL or language models (Sec~\ref{sec:exp_setup:models}), but may differ by their inputs and number of training examples (Sec~\ref{sec:exp_setup:eval_setup}). 

\subsection{Models}
\label{sec:exp_setup:models}

We experiment with two state-of-the-art VL models that can generate text conditioned on both text and images, as well as one language model. 

\paragraph{Open Flamingo.} Flamingo was initialized with a pre-trained LLM and a pre-trained vision model, and further trained on vision and language tasks, keeping the pre-trained models frozen. The interaction between the two modalities is facilitated with a gated cross-attention dense block. Since the original model is not publicly available, we use the open version, OpenFlamingo-9B \cite{open-flamingo}. OpenFlamingo is built on top of LLaMA 7B \cite{llama} and CLIP ViT/L-14 \cite{clip}, and was trained on 5M samples from the Multimodal C4 dataset \cite{multimodal-c4} and 10M samples from LAION-2B \cite{laion}.

\paragraph{MiniGPT4.} MiniGPT4 \cite{minigpt4} is similarly composed of frozen pre-trained language and vision models, and it employs a single projection layer to align the visual and language features. Since GPT4's architecture and training data remain a mystery, we utilize MiniGPT4 as an alternative to GPT4 \cite{gpt4}.\footnote{The version of GPT-4 available through the OpenAI API doesn't support images.} It has similar capabilities to GPT-4 in understanding and generating the context \cite{minigpt4}. For its language model, MiniGPT4 uses Vicuna \cite{vicuna}, which is built on top of LLaMA-13B and performs on par with ChatGPT \cite{gpt4}. For its vision component, it uses BLIP-2 \cite{blip2}, which consists of CLIP ViT-G/14 and a Q-Former architecture. MiniGPT4 was trained on various multimodal datasets, including images from LAION \cite{laion}, Conceptual Captions \cite{sharma-etal-2018-conceptual}, and SBU \cite{sbu-captions}. 

\paragraph{LLaMA} LLaMA \cite{llama} is a transformer-based language model that was trained on trillions of tokens from exclusively publicly-available data. The LLaMA-13B model outperforms GPT-3 \cite{gpt3} on most benchmarks. We use the LLaMA-7B model, which achieves comparable performance to the LLaMA-13B model on most benchmarks. Since LLaMA is a language model rather than a VL model, its access to the visual content is through the image caption and the OCR text alone.

\subsection{Evaluation Setup}
\label{sec:exp_setup:eval_setup}

\paragraph{Inputs.} We test the models with different input settings. In the setup which is the most comparable to humans, we provide the models with the meme and \textcolor{Purple}{title}. We also experiment with setups that aid the model. One such input is the \textcolor{Green}{image caption}, which can help the model focus on the language modality and ignore the image. The second such input is the \textcolor{Orange}{text inside the meme}, that we extracted using  EasyOCR,\footnote{\url{https://github.com/JaidedAI/EasyOCR}} which helps the model focus on the visual aspects of the image and includes the text inside the image as part of the language input. We incrementally added each of these inputs.

\paragraph{Learning Setups.} We evaluate all models in a zero-shot setup. Flamingo and LLaMA enable in-context learning, so we experiment with 4, 8, and 12 shots. An example prompt (including the meme, title, image caption, and text inside the meme) is illustrated in Figure~\ref{fig:few_shot}. MiniGPT4 works in a chat format, so rather than in-context learning, we use it in either a zero-shot setup, or fine-tuned on our training set. 

Lastly, motivated by \citet{visual-metaphors} and \newcite{multicot}, we also tested  models in a Chain of Thought (CoT) style prompting \cite{cot}. In our case, we elicit multi-step reasoning from the LLM by providing the \textcolor{Blue}{visual metaphors}, using the following prompt:
\begin{framed}
\small \noindent \tt
<image>This is a meme with the title ``\{\textcolor{Purple}{ title}\}''. \\
The image description is ``\{\textcolor{Green}{image caption}\}''. \\
The following text is written inside the meme: ``\{\textcolor{Orange}{OCR text}\}''. \\
What is the meme poster trying to convey? \\
Rationale: ``\textcolor{Blue}{\{keyword1\}'' is a metaphor for ``\{meaning1\}''. ``\{keyword2\}'' is a metaphor for ``\{meaning2\}}''. \\
Answer:
\end{framed}

\section{Results}
\label{sec:results}
\begin{table*}[!t]
    \small
    \centering
    \begin{tabular}{lllrrrr}
    \toprule
  \textbf{Model} & \textbf{Setup} & \textbf{Inputs} & \textbf{BLEU-4} &  \textbf{ROUGE-L} & \textbf{BERT-F1}  \\ \midrule
  \multirow{10}{*}{\textbf{Flamingo}} & \multirow{4}{*}{\textbf{zero-shot}} 
  & meme+\textcolor{Purple}{title} & 19.36 & 31.51  & 65.69 \\ 
  & & meme+\textcolor{Green}{img cap} & 16.10  & 29.08  & 64.71 \\ 
  & & meme+\textcolor{Purple}{title}+\textcolor{Green}{img cap} & 19.61  & 30.92  & 65.51 \\ 
  & & meme+\textcolor{Purple}{title}+\textcolor{Green}{img cap}+\textcolor{Orange}{OCR text} & 19.31  & 32.51  & 66.84 \\ \hhline{~-----}
  & \textbf{zero-shot CoT} & meme+\textcolor{Purple}{title}+\textcolor{Green}{img cap}+\textcolor{Orange}{OCR text}+\textcolor{Blue}{rationale} & 2.49 & 15.89 & 58.23 \\ \hhline{~-----}
  & \multirow{4}{*}{\textbf{few-shot}} & 
  meme+\textcolor{Purple}{title} & 25.89  & 39.41  & 70.83 \\ 
  & & meme+\textcolor{Green}{img cap} & 26.96  & 39.53  & 70.91 \\ 
  & & meme+\textcolor{Purple}{title}+\textcolor{Green}{img cap} & 26.44  & 39.42  & 71.04 \\ 
  & & meme+\textcolor{Purple}{title}+\textcolor{Green}{img cap}+\textcolor{Orange}{OCR text} & 26.73  & \textbf{43.47}  & 73.86 \\ \hhline{~-----}
  & \textbf{few-shot CoT} & meme+\textcolor{Purple}{title}+\textcolor{Green}{img cap}+\textcolor{Orange}{OCR text}+\textcolor{Blue}{rationale} & \textbf{27.02} & 43.46 & 74.32 \\ \midrule
  \multirow{8}{*}{\textbf{MiniGPT4}} & \multirow{5}{*}{\textbf{zero-shot}} & meme & 06.17  & 22.20  & 63.31 \\
  & & meme+\textcolor{Purple}{title} & 14.37  & 30.70  & 66.19 \\ 
  & & meme+\textcolor{Green}{img cap} & 10.36  & 26.22  & 64.39 \\ 
  & & meme+\textcolor{Purple}{title}+\textcolor{Green}{img cap} & 12.49  & 28.51  & 65.81 \\ 
  & & meme+\textcolor{Purple}{title}+\textcolor{Green}{img cap}+\textcolor{Orange}{OCR text} & 12.46  & 31.44  & 68.62 \\ \hhline{~-----}
  & \textbf{zero-shot CoT} & meme+\textcolor{Purple}{title}+\textcolor{Green}{img cap}+\textcolor{Orange}{OCR text}+\textcolor{Blue}{rationale} & 12.57  & 31.70  & 68.45 \\ \hhline{~-----}
  & \textbf{fine-tuned} & meme+\textcolor{Purple}{title}+\textcolor{Green}{img cap}+\textcolor{Orange}{OCR text} & 7.50  & 27.88  & 65.47 \\ \hhline{~-----} 
  & \textbf{fine-tuned CoT} & meme+\textcolor{Purple}{title}+\textcolor{Green}{img cap}+\textcolor{Orange}{OCR text}+\textcolor{Blue}{rationale} & 7.25  & 26.68  & 65.86 \\ \midrule
  \multirow{6}{*}{\textbf{LLaMA}} & \multirow{2}{*}{\textbf{zero-shot}} & \textcolor{Purple}{title}+\textcolor{Green}{img cap} & 19.72  & 31.42  & 66.38 \\  
  & & \textcolor{Purple}{title}+\textcolor{Green}{img cap}+\textcolor{Orange}{OCR text} & 20.77  & 36.48  & 69.67 \\ \hhline{~-----}
  & \textbf{zero-shot CoT} & \textcolor{Purple}{title}+\textcolor{Green}{img cap}+\textcolor{Orange}{OCR text}+\textcolor{Blue}{rationale} & 6.72 & 20.56 & 61.38 \\ \hhline{~-----} 
  & \multirow{2}{*}{\textbf{few-shot}} & \textcolor{Purple}{title}+\textcolor{Green}{img cap} & 26.41  & 38.70  & 70.01 \\ 
  & & \textcolor{Purple}{title}+\textcolor{Green}{img cap}+\textcolor{Orange}{OCR text} & 26.63  & 43.41  & \textbf{74.71} \\ \hhline{~-----} 
  & \textbf{few-shot CoT} & \textcolor{Purple}{title}+\textcolor{Green}{img cap}+\textcolor{Orange}{OCR text}+\textcolor{Blue}{rationale} & 26.40 & 42.95 & 74.00 \\ 
  \bottomrule
    \end{tabular}
    \caption{Performance in terms of automatic metrics of the various models and learning setups (with 4 shots for the few-shot setup). We report the full experimental results, including 8 shots and 12 shots, in Appendix~\ref{appendix:exp-results}.}
    \label{tab:auto_results}
\end{table*}

We evaluated the performance of the various models with both automatic metrics (Sec~\ref{sec:results:auto_eval}) and human evaluation (Sec~\ref{sec:results:human_eval}). We show that the vision and language modalities are complementary through ablation tests (Sec~\ref{sec:results:ablation}).

\subsection{Automatic Evaluation}
\label{sec:results:auto_eval}

To evaluate the quality of the generated captions, we use standard metrics for automatic evaluation of generative tasks: BLEU \cite{papineni-etal-2002-bleu} ROUGE \cite{lin-2004-rouge}, and BERTScore \cite{bertscore} (using \texttt{microsoft/deberta-xlarge-mnli}). BLEU and ROUGE are based on n-gram overlap between the generated captions and human-written reference captions, while BERTScore measures the semantic similarities between the two. 

Table~\ref{tab:auto_results} shows the performance of the various models and input setups in terms of these metrics. For the few-shot setup, we show the best performance across (4, 8, and 12 shots). See Appendix~\ref{appendix:exp-results} for the full results. 

\paragraph{Models.} Flamingo dominates MiniGPT4 across all metrics, with a gap of 15, 12, and 6 points in BLEU, ROUGE, and BertScore respectively for the best setups. This is likely due to the lengthy captions generated by MiniGPT4, despite the prompt including the instruction to generate a single sentence. Finally, the LLaMA model is highly competitive with Flamingo despite not having access to the image itself. It appears that the image captions and OCR text provide sufficient information.  

\paragraph{Learning Setups.} The Flamingo performance significantly improves from the zero-shot to few-shot setting, and continues to improve from 4 to 8 shots but slightly decreases at 12 shots (see Appendix~\ref{appendix:exp-results}). MiniGPT4 achieved better performance in the zero-shot setup, while fine-tuning its last layer significantly decrease the performance. As we show in Sec~\ref{sec:results:human_eval}, while the fine-tuned model learns to generate short captions, it tends to hallucinate more. We hypothesize that fine-tuning only the last layer is ineffective. 

\paragraph{Inputs.} In the few-shot setups, the best performance is achieved with as many of the inputs as possible, i.e. including both the image caption and the OCR text, despite the redundancy with the visual inputs. This might be due to suboptimal cross-modal interaction in VL models. While prior work showed that explicitly stating the metaphors helps image generation models generate better images \cite{visual-metaphors}, we did not see a similar gain in meme captioning.

\begin{figure*}[!t]
\centering\includegraphics[width=\textwidth]{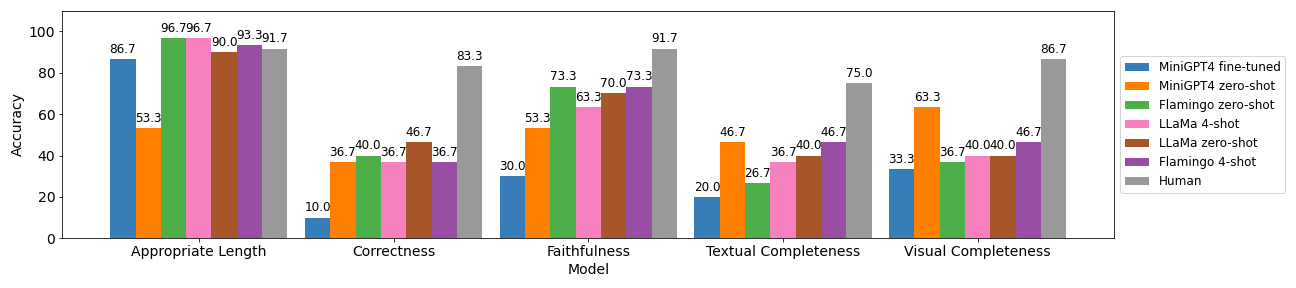}
    \caption{Performance in terms of human evaluation.}
    \label{fig:human-eval}
\end{figure*}

\subsection{Human Evaluation}
\label{sec:results:human_eval}

We focused on the models with the full set of inputs except for the rationales (meme+\textcolor{Purple}{title}+\textcolor{Green}{img cap}+\textcolor{Orange}{OCR text}) and evaluated the performance of all models (focusing on 4-shots for the few-shot setups), with respect to the following criteria: 

\begin{itemize}[itemjoin={{\hspace{0.2em}}}, itemsep=0.2em,leftmargin=*]
  \item \textbf{Correctness}: Does the caption correctly convey the meaning the meme poster wanted to convey?
  \item \textbf{Appropriate Length}: Is the caption length appropriate for conveying the meaning (i.e. it is not too verbose)?
  \item \textbf{Visual Completeness}: Does the caption describe all the important elements in the image?
  \item \textbf{Textual Completeness}: Does the caption describe all the important elements in the text inside the meme and the title text?
  \item \textbf{Faithfulness}: Are all the elements of the caption supported by either the visual or text elements (i.e. there are no made-up elements)?
\end{itemize}

\begin{figure*}[!t]
\centering\includegraphics[width=\textwidth]{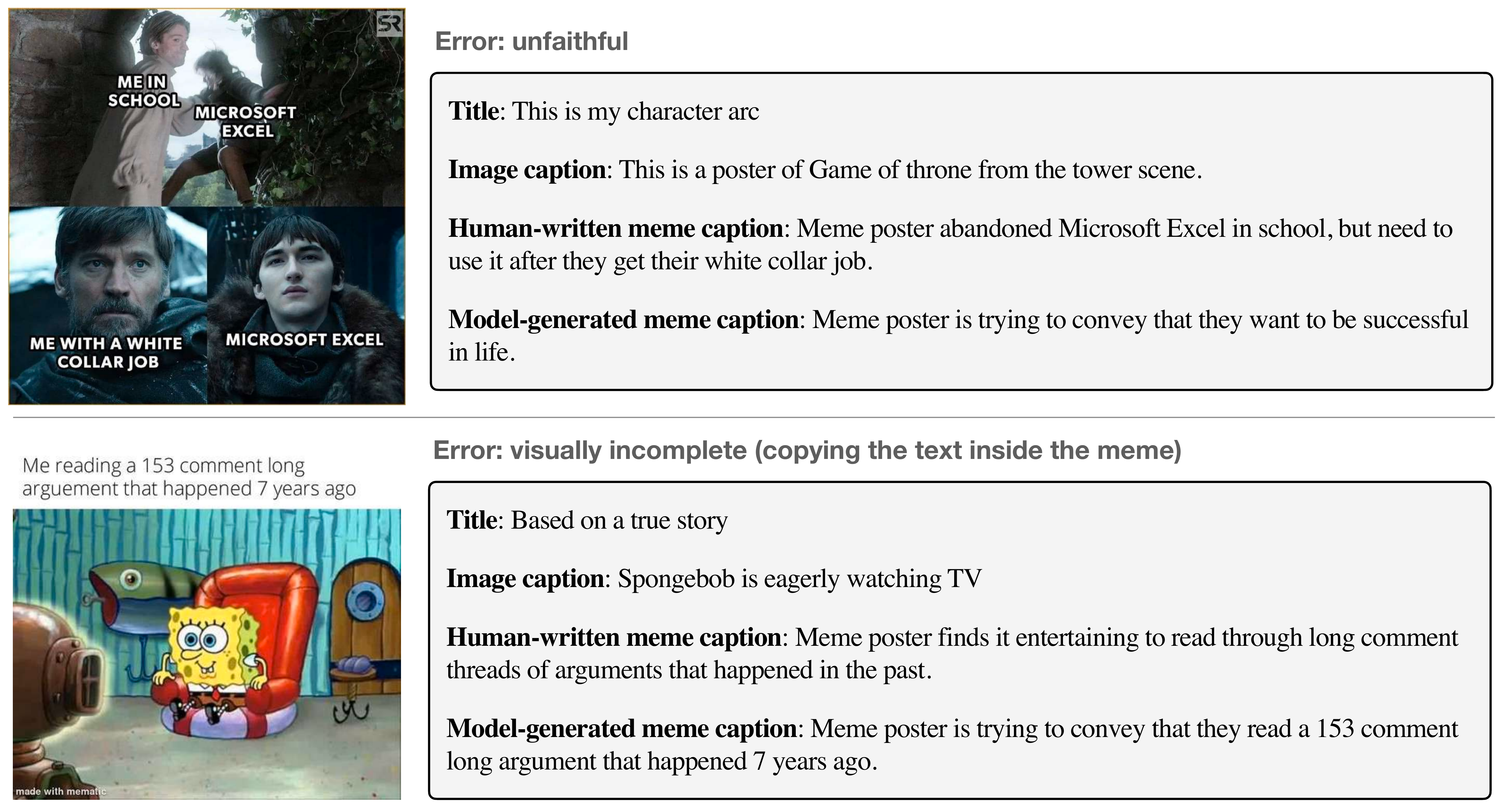}
    \caption{Examples of incorrect meme captions generated by the few-shot Flamingo model.}
    \label{fig:error-analysis-ex}
\end{figure*}

We randomly sampled 30 memes along with their model-generated and human-written captions. The annotation was performed by students in the lab. Figure~\ref{fig:human-eval} shows the performance according to the human evaluation. All models perform significantly worse than humans, except for appropriate length criteria, with 36.6, 29.3, 24.5, and 18.4 point differences on correctness, textual completeness, visual completeness, and faithfulness respectively. 

\paragraph{Models.} Model performance differs by criteria. Flamingo and LLaMA are more correct and faithful, while MiniGPT4 is more visually complete. 

\paragraph{Learning Setups.} For Flamingo, the few-shot models improve in textual and visual completeness upon the zero-shot model, but not in terms of correctness and faithfulness. This may suggest that while access to examples improves the model's understanding of the task, it might also confuse it with information irrelevant to the target meme. LLaMA doesn't gain any performance improvements from in-context examples, likely for the same reason. Without the visual features, it might struggle even more to separate the text (title, image caption, and OCR) of the different examples.

MiniGPT4 zero-shot is very verbose, but the fine-tuned model learns to output captions in the length of its training examples. Unfortunately, these captions are far worse than those of the zero-shot model in all criteria. We hypothesize that the frozen language and vision model may not have enough information about memes, and simply fine-tuning the last projection layer of the model is not enough to produce high-quality captions. This conclusion is consistent with \newcite{lima}, according to which most knowledge in LLM is learned during the pre-training stage.

\paragraph{Common Errors.} Figure~\ref{fig:error-analysis-ex} shows two examples of meme captions generated by Flamingo 4-shot along with the types of errors they exhibit. The top example demonstrates an unfaithful caption because neither the meme nor the title conveys anything about being successful in life. The bottom example illustrates a common error in which the model copies text from inside the meme while ignoring important visual elements. In this case, Spongebob's smile indicates the meme poster's positive attitude towards reading old and long forum threads, but the model-generated caption misses it. Another common error (not illustrated here) occurs when the model treats visual elements too literally, failing to interpret the metaphor. Finally, in some cases, the model might lack sufficient background knowledge to correctly interpret the meme. 

\begin{table}[!t]
    \small
    \centering
    \begin{tabular}{lllrrrr}
    \toprule
  \textbf{Model} & \textbf{k} & \textbf{Inputs} & $\Delta$\textbf{BL} &  $\Delta$\textbf{RG} & $\Delta$\textbf{BT}  \\ \midrule
  \multirow{6}{*}{\textbf{Flamingo}} &  \multirow{3}{*}{\textbf{0}} & full & 19.36  & 31.51  & 65.69 \\ 
  & & -title & -2.29 & -1.35 & -0.6 \\ 
  & & -meme & -1.49 & -1.93 & -1.71 \\ 
   \hhline{~-----}
  & \multirow{3}{*}{\textbf{4}} & full & 25.89  & 39.41 & 70.83 \\ 
  & & -title & +0.35 & +0.12 & -0.19 \\ 
  & & -meme & -0.14 & -0.85 & -1.86 \\ \midrule

  \multirow{3}{*}{\textbf{MiniGPT4}} &  \multirow{3}{*}{\textbf{0}} & full & 14.37  & 30.70  & 66.19 \\ 
  & & -title & -8.2 & -8.5 & -2.88 \\ 
  & & -meme & +3.5 & -1.12 & -2.21 \\ \midrule
\multirow{6}{*}{\textbf{LLaMA}} &  \multirow{3}{*}{\textbf{0}} & full & 19.72  & 31.42  & 66.38 \\ 
  & & -title & -0.88 & -0.93  & -0.62 \\
  & & -img cap & -1.85  & -1.84  & -2.4 \\
   \hhline{~-----}
  & \multirow{3}{*}{\textbf{4}} & full & 26.41  & 38.70  & 70.01 \\ 
  & & -title & -0.69 & -0.73  & -0.67 \\
  & & -img cap & -0.66  & -0.14  & -1.04 \\ 
   \bottomrule
    \end{tabular}
    \caption{Comparison models with both language and visual inputs (\textcolor{Purple}{title}+\textcolor{Green}{ima cap} for LLaMA, \textcolor{Purple}{title}+meme for VL models), compared to one modality. BL = BLEU, RG = ROUGE, BT = BERT. k = number of shots.}
    \label{tab:ablations}
\end{table}

\subsection{Ablation Tests}
\label{sec:results:ablation}

The analysis in Sec~\ref{sec:analysis:metaphor_types} shows that interpreting most memes in \dataset{} will require understanding both the visual and text modalities. We are interested in the extent that models make use of each modality. To that end, we perform an ablation test to exclude each modality. Table~\ref{tab:ablations} presents the results in terms of automatic metrics. 

In most cases, the best performance is achieved with both modalities. For Flamingo (zero-shot  and few-shot), excluding the meme results in more decrease in performance than excluding the title, indicating that the model relies more on the visual modality than the information provided by the title. The same is true for LLaMA (in both settings), for which excluding the image caption yields worse performance. This is expected since the title is typically secondary in informativeness to the meme. In addition, Flamingo still has access to the text inside the meme via visual features. 

Conversely, MiniGPT4 exhibits a higher dependency on textual modality, resulting in a significant decrease when the title is not provided. Since MiniGPT4 shows higher textual and visual completeness when the OCR text is provided (\S\ref{sec:results:human_eval}), we hypothesize that MiniGPT4 makes limited usage of the visual modality.

\section{Conclusion}
\label{sec:conclusion}
We present \dataset{}, the first meme captioning dataset. \dataset{} is challenging for the existing VL models, as it requires recognizing and interpreting visual metaphors, and ignoring the literal visual elements. The experimental results using state-of-the-art VL models indeed show that such models  are still far from human performance. In particular, they tend to treat visual elements too literally and copy text from inside the meme. Our work opens up interesting future research on recognizing visual metaphors, interpreting them with respect to a textual context, and generating meme captions that are complete with respect to both modalities without creating fake elements.

\section*{Limitations}
\label{sec:limitations}

\paragraph{Quality of Metaphor Annotations.} We put our best efforts into manually verifying the collected data, and indeed the human performance in Section~\ref{sec:results:human_eval} shows the human-written captions are of high quality. With that said, we noticed that the quality of the visual metaphors is inconsistent. We believe that while people are capable of explaining a meme, they don't always know to map the visual vehicles into textual targets. This likely explains why adding the metaphors as inputs didn't improve the performance. 

\paragraph{Subjectivity and Background Knowledge.} The meme captioning task involves employing background knowledge which may vary between annotators. To that end, we manually checked the meme captions to minimize the number of incorrect captions in the dataset. In addition, there is some level of subjectivity with respect to the evaluation criteria for the meme caption quality. For this reason, we ensured a high quality of annotations by having in-house annotators that could ask clarification questions, but some subjectivity still remains. 


\section*{Ethics Statement}
\label{sec:ethics}
\paragraph{Data}
All the datasets used in our work are publicly available. Our dataset is collected from Reddit and may contain offensive, hateful, or sexual content. Despite our best efforts to filter them out as described in Section \ref{sec:dataset}, we found people have different criteria for what they perceive as offensive, hateful, or sexual, and thus, such content may still exist in our data.

\paragraph{Data Collection} We use Amazon Mechanical Turk to collect 6.3K image descriptions and 7.7K meme captions. The annotators were compensated with an average hourly wage of \$13, which is comparable to the US minimum wage. We did not collect any personal information from annotators.

\paragraph{Models} Our dataset may include some offensive content or mild expletives and this can amplify potentially biased and unethical answers. In addition, the large pre-trained VL models we used for the experiments are trained on a large-scale publicly available web corpus and may also bring some bias when generating sentences. 

\section*{Acknowledgements}
This work was funded, in part, by the Vector Institute for AI, Canada CIFAR AI Chairs program, an NSERC discovery grant, and a research gift from AI2. 

\bibliography{anthology,custom}

\begin{thebibliography}{41}
\expandafter\ifx\csname natexlab\endcsname\relax\def\natexlab#1{#1}\fi

\bibitem[{Aghazadeh et~al.(2022)Aghazadeh, Fayyaz, and
  Yaghoobzadeh}]{aghazadeh-etal-2022-metaphors}
Ehsan Aghazadeh, Mohsen Fayyaz, and Yadollah Yaghoobzadeh. 2022.
\newblock \href {https://doi.org/10.18653/v1/2022.acl-long.144} {Metaphors in
  pre-trained language models: Probing and generalization across datasets and
  languages}.
\newblock In \emph{Proceedings of the 60th Annual Meeting of the Association
  for Computational Linguistics (Volume 1: Long Papers)}, pages 2037--2050,
  Dublin, Ireland. Association for Computational Linguistics.

\bibitem[{Alayrac et~al.(2022)Alayrac, Donahue, Luc, Miech, Barr, Hasson, Lenc,
  Mensch, Millican, Reynolds, Ring, Rutherford, Cabi, Han, Gong, Samangooei,
  Monteiro, Menick, Borgeaud, Brock, Nematzadeh, Sharifzadeh, Binkowski,
  Barreira, Vinyals, Zisserman, and Simonyan}]{flamingo}
Jean-Baptiste Alayrac, Jeff Donahue, Pauline Luc, Antoine Miech, Iain Barr,
  Yana Hasson, Karel Lenc, Arthur Mensch, Katherine Millican, Malcolm Reynolds,
  Roman Ring, Eliza Rutherford, Serkan Cabi, Tengda Han, Zhitao Gong, Sina
  Samangooei, Marianne Monteiro, Jacob Menick, Sebastian Borgeaud, Andrew
  Brock, Aida Nematzadeh, Sahand Sharifzadeh, Mikolaj Binkowski, Ricardo
  Barreira, Oriol Vinyals, Andrew Zisserman, and Karen Simonyan. 2022.
\newblock \href {https://openreview.net/forum?id=EbMuimAbPbs} {Flamingo: a
  visual language model for few-shot learning}.
\newblock In \emph{Advances in Neural Information Processing Systems}.

\bibitem[{Alikhani et~al.(2020)Alikhani, Sharma, Li, Soricut, and
  Stone}]{alikhani-etal-2020-cross}
Malihe Alikhani, Piyush Sharma, Shengjie Li, Radu Soricut, and Matthew Stone.
  2020.
\newblock \href {https://doi.org/10.18653/v1/2020.acl-main.583} {Cross-modal
  coherence modeling for caption generation}.
\newblock In \emph{Proceedings of the 58th Annual Meeting of the Association
  for Computational Linguistics}, pages 6525--6535, Online. Association for
  Computational Linguistics.

\bibitem[{Awadalla et~al.(2023)Awadalla, Gao, Gardner, Hessel, Hanafy, Zhu,
  Marathe, Bitton, Gadre, Jitsev, Kornblith, Koh, Ilharco, Wortsman, and
  Schmidt}]{open-flamingo}
Anas Awadalla, Irena Gao, Joshua Gardner, Jack Hessel, Yusuf Hanafy, Wanrong
  Zhu, Kalyani Marathe, Yonatan Bitton, Samir Gadre, Jenia Jitsev, Simon
  Kornblith, Pang~Wei Koh, Gabriel Ilharco, Mitchell Wortsman, and Ludwig
  Schmidt. 2023.
\newblock \href {https://doi.org/10.5281/zenodo.7733589} {Openflamingo}.

\bibitem[{Bitton-Guetta et~al.(2023)Bitton-Guetta, Bitton, Hessel, Schmidt,
  Elovici, Stanovsky, and Schwartz}]{bittonguetta2023breaking}
Nitzan Bitton-Guetta, Yonatan Bitton, Jack Hessel, Ludwig Schmidt, Yuval
  Elovici, Gabriel Stanovsky, and Roy Schwartz. 2023.
\newblock \href {http://arxiv.org/abs/2303.07274} {Breaking common sense:
  Whoops! a vision-and-language benchmark of synthetic and compositional
  images}.

\bibitem[{Brown et~al.(2020)Brown, Mann, Ryder, Subbiah, Kaplan, Dhariwal,
  Neelakantan, Shyam, Sastry, Askell, Agarwal, Herbert-Voss, Krueger, Henighan,
  Child, Ramesh, Ziegler, Wu, Winter, Hesse, Chen, Sigler, Litwin, Gray, Chess,
  Clark, Berner, McCandlish, Radford, Sutskever, and Amodei}]{gpt3}
Tom~B. Brown, Benjamin Mann, Nick Ryder, Melanie Subbiah, Jared Kaplan,
  Prafulla Dhariwal, Arvind Neelakantan, Pranav Shyam, Girish Sastry, Amanda
  Askell, Sandhini Agarwal, Ariel Herbert-Voss, Gretchen Krueger, Tom Henighan,
  Rewon Child, Aditya Ramesh, Daniel~M. Ziegler, Jeffrey Wu, Clemens Winter,
  Christopher Hesse, Mark Chen, Eric Sigler, Mateusz Litwin, Scott Gray,
  Benjamin Chess, Jack Clark, Christopher Berner, Sam McCandlish, Alec Radford,
  Ilya Sutskever, and Dario Amodei. 2020.
\newblock \href {http://arxiv.org/abs/2005.14165} {Language models are few-shot
  learners}.

\bibitem[{Buchel(2012)}]{buchel2012internet}
Branislav Buchel. 2012.
\newblock Internet memes as means of communication.
\newblock \emph{Brno: Masaryk University}.

\bibitem[{Chakrabarty et~al.(2022)Chakrabarty, Choi, and
  Shwartz}]{chakrabarty-etal-2022-rocket}
Tuhin Chakrabarty, Yejin Choi, and Vered Shwartz. 2022.
\newblock \href {https://doi.org/10.1162/tacl_a_00478} {It{'}s not rocket
  science: Interpreting figurative language in narratives}.
\newblock \emph{Transactions of the Association for Computational Linguistics},
  10:589--606.

\bibitem[{Chakrabarty et~al.(2021{\natexlab{a}})Chakrabarty, Ghosh, Poliak, and
  Muresan}]{chakrabarty-etal-2021-figurative}
Tuhin Chakrabarty, Debanjan Ghosh, Adam Poliak, and Smaranda Muresan.
  2021{\natexlab{a}}.
\newblock \href {https://doi.org/10.18653/v1/2021.findings-acl.297} {Figurative
  language in recognizing textual entailment}.
\newblock In \emph{Findings of the Association for Computational Linguistics:
  ACL-IJCNLP 2021}, pages 3354--3361, Online. Association for Computational
  Linguistics.

\bibitem[{Chakrabarty et~al.(2023)Chakrabarty, Saakyan, Winn, Panagopoulou,
  Yang, Apidianaki, and Muresan}]{visual-metaphors}
Tuhin Chakrabarty, Arkady Saakyan, Olivia Winn, Artemis Panagopoulou, Yue Yang,
  Marianna Apidianaki, and Smaranda Muresan. 2023.
\newblock I spy a metaphor: Large language models and diffusion models
  co-create visual metaphors.
\newblock In \emph{Findings of ACL}.

\bibitem[{Chakrabarty et~al.(2021{\natexlab{b}})Chakrabarty, Zhang, Muresan,
  and Peng}]{chakrabarty-etal-2021-mermaid}
Tuhin Chakrabarty, Xurui Zhang, Smaranda Muresan, and Nanyun Peng.
  2021{\natexlab{b}}.
\newblock \href {https://doi.org/10.18653/v1/2021.naacl-main.336} {{MERMAID}:
  Metaphor generation with symbolism and discriminative decoding}.
\newblock In \emph{Proceedings of the 2021 Conference of the North American
  Chapter of the Association for Computational Linguistics: Human Language
  Technologies}, pages 4250--4261, Online. Association for Computational
  Linguistics.

\bibitem[{Chiang et~al.(2023)Chiang, Li, Lin, Sheng, Wu, Zhang, Zheng, Zhuang,
  Zhuang, Gonzalez, Stoica, and Xing}]{vicuna}
Wei-Lin Chiang, Zhuohan Li, Zi~Lin, Ying Sheng, Zhanghao Wu, Hao Zhang, Lianmin
  Zheng, Siyuan Zhuang, Yonghao Zhuang, Joseph~E. Gonzalez, Ion Stoica, and
  Eric~P. Xing. 2023.
\newblock \href {https://vicuna.lmsys.org} {Vicuna: An open-source chatbot
  impressing gpt-4 with 90\% chatgpt quality}.

\bibitem[{Choi et~al.(2021)Choi, Lee, Choi, Park, Lee, Lee, and
  Lee}]{choi-etal-2021-melbert}
Minjin Choi, Sunkyung Lee, Eunseong Choi, Heesoo Park, Junhyuk Lee, Dongwon
  Lee, and Jongwuk Lee. 2021.
\newblock \href {https://doi.org/10.18653/v1/2021.naacl-main.141} {{M}el{BERT}:
  Metaphor detection via contextualized late interaction using metaphorical
  identification theories}.
\newblock In \emph{Proceedings of the 2021 Conference of the North American
  Chapter of the Association for Computational Linguistics: Human Language
  Technologies}, pages 1763--1773, Online. Association for Computational
  Linguistics.

\bibitem[{Dodge et~al.(2015)Dodge, Hong, and
  Stickles}]{dodge-etal-2015-metanet}
Ellen Dodge, Jisup Hong, and Elise Stickles. 2015.
\newblock \href {https://doi.org/10.3115/v1/W15-1405} {{M}eta{N}et: Deep
  semantic automatic metaphor analysis}.
\newblock In \emph{Proceedings of the Third Workshop on Metaphor in {NLP}},
  pages 40--49, Denver, Colorado. Association for Computational Linguistics.

\bibitem[{Forceville(1996)}]{forceville1996pictorial}
Charles Forceville. 1996.
\newblock \emph{Pictorial metaphor in advertising}.
\newblock Psychology Press.

\bibitem[{Kiela et~al.(2021)Kiela, Firooz, Mohan, Goswami, Singh, Fitzpatrick,
  Bull, Lipstein, Nelli, Zhu, Muennighoff, Velioglu, Rose, Lippe, Holla,
  Chandra, Rajamanickam, Antoniou, Shutova, Yannakoudakis, Sandulescu, Ozertem,
  Pantel, Specia, and Parikh}]{hateful-memes}
Douwe Kiela, Hamed Firooz, Aravind Mohan, Vedanuj Goswami, Amanpreet Singh,
  Casey~A. Fitzpatrick, Peter Bull, Greg Lipstein, Tony Nelli, Ron Zhu, Niklas
  Muennighoff, Riza Velioglu, Jewgeni Rose, Phillip Lippe, Nithin Holla,
  Shantanu Chandra, Santhosh Rajamanickam, Georgios Antoniou, Ekaterina
  Shutova, Helen Yannakoudakis, Vlad Sandulescu, Umut Ozertem, Patrick Pantel,
  Lucia Specia, and Devi Parikh. 2021.
\newblock \href {https://proceedings.mlr.press/v133/kiela21a.html} {The hateful
  memes challenge: Competition report}.
\newblock In \emph{Proceedings of the NeurIPS 2020 Competition and
  Demonstration Track}, volume 133 of \emph{Proceedings of Machine Learning
  Research}, pages 344--360. PMLR.

\bibitem[{Li et~al.(2023)Li, Li, Savarese, and Hoi}]{blip2}
Junnan Li, Dongxu Li, Silvio Savarese, and Steven Hoi. 2023.
\newblock \href {http://arxiv.org/abs/2301.12597} {Blip-2: Bootstrapping
  language-image pre-training with frozen image encoders and large language
  models}.

\bibitem[{Lin(2004)}]{lin-2004-rouge}
Chin-Yew Lin. 2004.
\newblock \href {https://aclanthology.org/W04-1013} {{ROUGE}: A package for
  automatic evaluation of summaries}.
\newblock In \emph{Text Summarization Branches Out}, pages 74--81, Barcelona,
  Spain. Association for Computational Linguistics.

\bibitem[{OpenAI(2023)}]{gpt4}
OpenAI. 2023.
\newblock \href {http://arxiv.org/abs/2303.08774} {Gpt-4 technical report}.

\bibitem[{Ordonez et~al.(2011)Ordonez, Kulkarni, and Berg}]{sbu-captions}
Vicente Ordonez, Girish Kulkarni, and Tamara Berg. 2011.
\newblock \href
  {https://proceedings.neurips.cc/paper/2011/file/5dd9db5e033da9c6fb5ba83c7a7ebea9-Paper.pdf}
  {Im2text: Describing images using 1 million captioned photographs}.
\newblock In \emph{Advances in Neural Information Processing Systems},
  volume~24. Curran Associates, Inc.

\bibitem[{Papineni et~al.(2002)Papineni, Roukos, Ward, and
  Zhu}]{papineni-etal-2002-bleu}
Kishore Papineni, Salim Roukos, Todd Ward, and Wei-Jing Zhu. 2002.
\newblock \href {https://doi.org/10.3115/1073083.1073135} {{B}leu: a method for
  automatic evaluation of machine translation}.
\newblock In \emph{Proceedings of the 40th Annual Meeting of the Association
  for Computational Linguistics}, pages 311--318, Philadelphia, Pennsylvania,
  USA. Association for Computational Linguistics.

\bibitem[{Qu et~al.(2022)Qu, Li, Zhao, Dev, and Chang}]{qu2022disinfomeme}
Jingnong Qu, Liunian~Harold Li, Jieyu Zhao, Sunipa Dev, and Kai-Wei Chang.
  2022.
\newblock \href {http://arxiv.org/abs/2205.12617} {Disinfomeme: A multimodal
  dataset for detecting meme intentionally spreading out disinformation}.

\bibitem[{Radford et~al.(2021)Radford, Kim, Hallacy, Ramesh, Goh, Agarwal,
  Sastry, Askell, Mishkin, Clark, Krueger, and Sutskever}]{clip}
Alec Radford, Jong~Wook Kim, Chris Hallacy, Aditya Ramesh, Gabriel Goh,
  Sandhini Agarwal, Girish Sastry, Amanda Askell, Pamela Mishkin, Jack Clark,
  Gretchen Krueger, and Ilya Sutskever. 2021.
\newblock \href {https://proceedings.mlr.press/v139/radford21a.html} {Learning
  transferable visual models from natural language supervision}.
\newblock In \emph{Proceedings of the 38th International Conference on Machine
  Learning}, volume 139 of \emph{Proceedings of Machine Learning Research},
  pages 8748--8763. PMLR.

\bibitem[{Schuhmann et~al.(2022)Schuhmann, Beaumont, Vencu, Gordon, Wightman,
  Cherti, Coombes, Katta, Mullis, Wortsman, Schramowski, Kundurthy, Crowson,
  Schmidt, Kaczmarczyk, and Jitsev}]{laion}
Christoph Schuhmann, Romain Beaumont, Richard Vencu, Cade Gordon, Ross
  Wightman, Mehdi Cherti, Theo Coombes, Aarush Katta, Clayton Mullis, Mitchell
  Wortsman, Patrick Schramowski, Srivatsa Kundurthy, Katherine Crowson, Ludwig
  Schmidt, Robert Kaczmarczyk, and Jenia Jitsev. 2022.
\newblock \href {http://arxiv.org/abs/2210.08402} {Laion-5b: An open
  large-scale dataset for training next generation image-text models}.

\bibitem[{Scott(2021)}]{scott2021memes}
Kate Scott. 2021.
\newblock Memes as multimodal metaphors: A relevance theory analysis.
\newblock \emph{Pragmatics \& Cognition}, 28(2):277--298.

\bibitem[{Sharma et~al.(2020)Sharma, Paka, Bhageria, Das, Poria, Chakraborty,
  and Gamb{\"a}ck}]{chhavi2020memotion}
Chhavi Sharma, William Paka, Scott, Deepesh Bhageria, Amitava Das, Soujanya
  Poria, Tanmoy Chakraborty, and Bj{\"o}rn Gamb{\"a}ck. 2020.
\newblock {Task Report: Memotion Analysis 1.0 @SemEval 2020: The Visuo-Lingual
  Metaphor!}
\newblock In \emph{Proceedings of the 14th International Workshop on Semantic
  Evaluation ({S}em{E}val-2020)}, Barcelona, Spain. Association for
  Computational Linguistics.

\bibitem[{Sharma et~al.(2018)Sharma, Ding, Goodman, and
  Soricut}]{sharma-etal-2018-conceptual}
Piyush Sharma, Nan Ding, Sebastian Goodman, and Radu Soricut. 2018.
\newblock \href {https://doi.org/10.18653/v1/P18-1238} {Conceptual captions: A
  cleaned, hypernymed, image alt-text dataset for automatic image captioning}.
\newblock In \emph{Proceedings of the 56th Annual Meeting of the Association
  for Computational Linguistics (Volume 1: Long Papers)}, pages 2556--2565,
  Melbourne, Australia. Association for Computational Linguistics.

\bibitem[{Sharma et~al.(2023)Sharma, Kulkarni, Suresh, Mathur, Nakov, Akhtar,
  and Chakraborty}]{sharma-etal-2023-characterizing}
Shivam Sharma, Atharva Kulkarni, Tharun Suresh, Himanshi Mathur, Preslav Nakov,
  Md.~Shad Akhtar, and Tanmoy Chakraborty. 2023.
\newblock \href {https://aclanthology.org/2023.eacl-main.157} {Characterizing
  the entities in harmful memes: Who is the hero, the villain, the victim?}
\newblock In \emph{Proceedings of the 17th Conference of the European Chapter
  of the Association for Computational Linguistics}, pages 2149--2163,
  Dubrovnik, Croatia. Association for Computational Linguistics.

\bibitem[{Stowe et~al.(2021)Stowe, Beck, and
  Gurevych}]{stowe-etal-2021-exploring}
Kevin Stowe, Nils Beck, and Iryna Gurevych. 2021.
\newblock \href {https://doi.org/10.18653/v1/2021.conll-1.26} {Exploring
  metaphoric paraphrase generation}.
\newblock In \emph{Proceedings of the 25th Conference on Computational Natural
  Language Learning}, pages 323--336, Online. Association for Computational
  Linguistics.

\bibitem[{Suvorov et~al.(2021)Suvorov, Logacheva, Mashikhin, Remizova, Ashukha,
  Silvestrov, Kong, Goka, Park, and Lempitsky}]{suvorov2021resolution}
Roman Suvorov, Elizaveta Logacheva, Anton Mashikhin, Anastasia Remizova,
  Arsenii Ashukha, Aleksei Silvestrov, Naejin Kong, Harshith Goka, Kiwoong
  Park, and Victor Lempitsky. 2021.
\newblock Resolution-robust large mask inpainting with fourier convolutions.
\newblock \emph{arXiv preprint arXiv:2109.07161}.

\bibitem[{Tanaka et~al.(2022)Tanaka, Yamane, Mori, Mukuta, and
  Harada}]{tanaka-etal-2022-learning}
Kohtaro Tanaka, Hiroaki Yamane, Yusuke Mori, Yusuke Mukuta, and Tatsuya Harada.
  2022.
\newblock \href {https://aclanthology.org/2022.cai-1.9} {Learning to evaluate
  humor in memes based on the incongruity theory}.
\newblock In \emph{Proceedings of the Second Workshop on When Creative AI Meets
  Conversational AI}, pages 81--93, Gyeongju, Republic of Korea. Association
  for Computational Linguistics.

\bibitem[{Touvron et~al.(2023)Touvron, Lavril, Izacard, Martinet, Lachaux,
  Lacroix, Rozière, Goyal, Hambro, Azhar, Rodriguez, Joulin, Grave, and
  Lample}]{llama}
Hugo Touvron, Thibaut Lavril, Gautier Izacard, Xavier Martinet, Marie-Anne
  Lachaux, Timothée Lacroix, Baptiste Rozière, Naman Goyal, Eric Hambro,
  Faisal Azhar, Aurelien Rodriguez, Armand Joulin, Edouard Grave, and Guillaume
  Lample. 2023.
\newblock \href {http://arxiv.org/abs/2302.13971} {Llama: Open and efficient
  foundation language models}.

\bibitem[{Wei et~al.(2022)Wei, Wang, Schuurmans, Bosma, ichter, Xia, Chi, Le,
  and Zhou}]{cot}
Jason Wei, Xuezhi Wang, Dale Schuurmans, Maarten Bosma, brian ichter, Fei Xia,
  Ed~Chi, Quoc~V Le, and Denny Zhou. 2022.
\newblock \href
  {https://proceedings.neurips.cc/paper_files/paper/2022/file/9d5609613524ecf4f15af0f7b31abca4-Paper-Conference.pdf}
  {Chain-of-thought prompting elicits reasoning in large language models}.
\newblock In \emph{Advances in Neural Information Processing Systems},
  volume~35, pages 24824--24837. Curran Associates, Inc.

\bibitem[{Xu et~al.(2022)Xu, Li, Zheng, Naseriparsa, Zhao, Lin, and
  Xia}]{xu2022met}
Bo~Xu, Tingting Li, Junzhe Zheng, Mehdi Naseriparsa, Zhehuan Zhao, Hongfei Lin,
  and Feng Xia. 2022.
\newblock Met-meme: A multimodal meme dataset rich in metaphors.
\newblock In \emph{Proceedings of the 45th International ACM SIGIR Conference
  on Research and Development in Information Retrieval}, pages 2887--2899.

\bibitem[{Zhang et~al.(2021)Zhang, Zhang, Zhang, Yang, and
  Lin}]{zhang-etal-2021-multimet}
Dongyu Zhang, Minghao Zhang, Heting Zhang, Liang Yang, and Hongfei Lin. 2021.
\newblock \href {https://doi.org/10.18653/v1/2021.acl-long.249} {{M}ulti{MET}:
  A multimodal dataset for metaphor understanding}.
\newblock In \emph{Proceedings of the 59th Annual Meeting of the Association
  for Computational Linguistics and the 11th International Joint Conference on
  Natural Language Processing (Volume 1: Long Papers)}, pages 3214--3225,
  Online. Association for Computational Linguistics.

\bibitem[{Zhang et~al.(2022)Zhang, Roller, Goyal, Artetxe, Chen, Chen, Dewan,
  Diab, Li, Lin, Mihaylov, Ott, Shleifer, Shuster, Simig, Koura, Sridhar, Wang,
  and Zettlemoyer}]{opt}
Susan Zhang, Stephen Roller, Naman Goyal, Mikel Artetxe, Moya Chen, Shuohui
  Chen, Christopher Dewan, Mona Diab, Xian Li, Xi~Victoria Lin, Todor Mihaylov,
  Myle Ott, Sam Shleifer, Kurt Shuster, Daniel Simig, Punit~Singh Koura, Anjali
  Sridhar, Tianlu Wang, and Luke Zettlemoyer. 2022.
\newblock \href {http://arxiv.org/abs/2205.01068} {Opt: Open pre-trained
  transformer language models}.

\bibitem[{Zhang et~al.(2020)Zhang, Kishore, Wu, Weinberger, and
  Artzi}]{bertscore}
Tianyi Zhang, Varsha Kishore, Felix Wu, Kilian~Q. Weinberger, and Yoav Artzi.
  2020.
\newblock \href {http://arxiv.org/abs/1904.09675} {Bertscore: Evaluating text
  generation with bert}.

\bibitem[{Zhang et~al.(2023)Zhang, Zhang, Li, Zhao, Karypis, and
  Smola}]{multicot}
Zhuosheng Zhang, Aston Zhang, Mu~Li, Hai Zhao, George Karypis, and Alex Smola.
  2023.
\newblock Multimodal chain-of-thought reasoning in language models.
\newblock \emph{arXiv preprint arXiv:2302.00923}.

\bibitem[{Zhou et~al.(2023)Zhou, Liu, Xu, Iyer, Sun, Mao, Ma, Efrat, Yu, Yu,
  Zhang, Ghosh, Lewis, Zettlemoyer, and Levy}]{lima}
Chunting Zhou, Pengfei Liu, Puxin Xu, Srini Iyer, Jiao Sun, Yuning Mao, Xuezhe
  Ma, Avia Efrat, Ping Yu, Lili Yu, Susan Zhang, Gargi Ghosh, Mike Lewis, Luke
  Zettlemoyer, and Omer Levy. 2023.
\newblock \href {http://arxiv.org/abs/2305.11206} {Lima: Less is more for
  alignment}.

\bibitem[{Zhu et~al.(2023{\natexlab{a}})Zhu, Chen, Shen, Li, and
  Elhoseiny}]{minigpt4}
Deyao Zhu, Jun Chen, Xiaoqian Shen, Xiang Li, and Mohamed Elhoseiny.
  2023{\natexlab{a}}.
\newblock \href {http://arxiv.org/abs/2304.10592} {Minigpt-4: Enhancing
  vision-language understanding with advanced large language models}.

\bibitem[{Zhu et~al.(2023{\natexlab{b}})Zhu, Hessel, Awadalla, Gadre, Dodge,
  Fang, Yu, Schmidt, Wang, and Choi}]{multimodal-c4}
Wanrong Zhu, Jack Hessel, Anas Awadalla, Samir~Yitzhak Gadre, Jesse Dodge, Alex
  Fang, Youngjae Yu, Ludwig Schmidt, William~Yang Wang, and Yejin Choi.
  2023{\natexlab{b}}.
\newblock \href {http://arxiv.org/abs/2304.06939} {Multimodal c4: An open,
  billion-scale corpus of images interleaved with text}.

\end{thebibliography}
\bibliographystyle{acl_natbib}

\appendix
\section{Additional Experimental Results}
\label{appendix:exp-results}

We show the full experimental results in Table~\ref{tab:full-result}.

\begin{table*}[!t]
    \small
    \centering
    \begin{tabular}{lllcccc}
    \toprule
  \textbf{Model} & \# \textbf{Shots} & \textbf{Input} & \textbf{BLEU-4} &  \textbf{ROUGE-L} & \textbf{BERT-F1}  \\ \midrule
  \multirow{24}{*}{\textbf{Flamingo}} & \multirow{5}{*}{\textbf{0-shot}} & meme & 17.07  & 30.16  & 65.09 \\
  & & meme+\textcolor{Purple}{title} & 19.36 & 31.51  & 65.69 \\ 
  & & meme+\textcolor{Green}{img cap} & 16.10  & 29.08  & 64.71 \\ 
  & & meme+\textcolor{Purple}{title}+\textcolor{Green}{img cap} & 19.61  & 30.92  & 65.51 \\ 
  & & meme+\textcolor{Purple}{title}+\textcolor{Green}{img cap}+\textcolor{Orange}{OCR text} & 19.31  & 32.51  & 66.84 \\ \hhline{~-----}
  & \textbf{0-shot CoT} & meme+\textcolor{Purple}{title}+\textcolor{Green}{img cap}+\textcolor{Orange}{OCR text}+\textcolor{Blue}{rationale} & 2.49 & 15.89 & 58.23 \\ \hhline{~-----}
  & \multirow{5}{*}{\textbf{4-shot}} & meme & 26.24  & 39.53  & 70.62 \\
  & & meme+\textcolor{Purple}{title} & 25.89  & 39.41  & 70.83 \\ 
  & & meme+\textcolor{Green}{img cap} & 26.96  & 39.53  & 70.91 \\ 
  & & meme+\textcolor{Purple}{title}+\textcolor{Green}{img cap} & 26.44  & 39.42  & 71.04 \\ 
  & & meme+\textcolor{Purple}{title}+\textcolor{Green}{img cap}+\textcolor{Orange}{OCR text} & 26.73  & 43.47  & 73.86 \\ \hhline{~-----}
  & \textbf{4-shot CoT} & meme+\textcolor{Purple}{title}+\textcolor{Green}{img cap}+\textcolor{Orange}{OCR text}+\textcolor{Blue}{rationale} & 27.02 & 43.46 & 74.32 \\ \hhline{~-----}
  & \multirow{5}{*}{\textbf{8-shot}} & meme & 27.38  & 39.96  & 70.92 \\
  & & meme+\textcolor{Purple}{title} & 26.99  & 40.00  & 71.26 \\ 
  & & meme+\textcolor{Green}{img cap} & 28.11  & 40.32  & 71.24 \\ 
  & & meme+\textcolor{Purple}{title}+\textcolor{Green}{img cap} & 27.30  & 40.00  & 71.32 \\ 
  & & meme+\textcolor{Purple}{title}+\textcolor{Green}{img cap}+\textcolor{Orange}{OCR text} & 28.70  & 43.54  & 74.33 \\ \hhline{~-----}
  & \textbf{8-shot CoT} & meme+\textcolor{Purple}{title}+\textcolor{Green}{img cap}+\textcolor{Orange}{OCR text}+\textcolor{Blue}{rationale} & - & - & - \\ \hhline{~-----}
  & \multirow{5}{*}{\textbf{12-shot}} & meme & 26.74  & 38.89  & 70.20 \\
  & & meme+\textcolor{Purple}{title} & 27.32  & 40.13  & 70.86 \\ 
  & & meme+\textcolor{Green}{img cap} & 26.63  & 39.24  & 70.49 \\ 
  & & meme+\textcolor{Purple}{title}+\textcolor{Green}{img cap} & 27.09  & 39.60  & 70.48 \\ 
  & & meme+\textcolor{Purple}{title}+\textcolor{Green}{img cap}+\textcolor{Orange}{OCR text} & -  & -  & - \\ \hhline{~-----}
  & \textbf{12-shot CoT} & meme+\textcolor{Purple}{title}+\textcolor{Green}{img cap}+\textcolor{Orange}{OCR text}+\textcolor{Blue}{rationale} & - & - & - \\ \midrule
  \multirow{20}{*}{\textbf{LLaMA}} & \multirow{4}{*}{\textbf{0-shot}} & \textcolor{Purple}{title} & 17.87  & 29.58  & 63.98 \\
  & & \textcolor{Green}{img cap} & 18.84  & 30.49  & 65.76 \\
  & & \textcolor{Purple}{title}+\textcolor{Green}{img cap} & 19.72  & 31.42  & 66.38 \\ 
  & & \textcolor{Purple}{title}+\textcolor{Green}{img cap}+\textcolor{Orange}{OCR text} & 20.77  & 36.48  & 69.67 \\ \hhline{~-----}
  & \textbf{0-shot CoT} & \textcolor{Purple}{title}+\textcolor{Green}{img cap}+\textcolor{Orange}{OCR text}+\textcolor{Blue}{rationale} & 6.72 & 20.56 & 61.38 \\ \hhline{~-----}
  & \multirow{4}{*}{\textbf{4-shot}} & \textcolor{Purple}{title} & 25.75  & 38.56  & 68.97 \\
  & & \textcolor{Green}{img cap} & 25.72  & 37.97  & 69.34 \\
  & & \textcolor{Purple}{title}+\textcolor{Green}{img cap} & 26.41  & 38.70  & 70.01 \\ 
  & & \textcolor{Purple}{title}+\textcolor{Green}{img cap}+\textcolor{Orange}{OCR text} & 26.63  & 43.41  & 74.71 \\ \hhline{~-----}
  & \textbf{4-shot CoT} & \textcolor{Purple}{title}+\textcolor{Green}{img cap}+\textcolor{Orange}{OCR text}+\textcolor{Blue}{rationale} & 26.40 & 42.95 & 74.00 \\ \hhline{~-----}
  & \multirow{4}{*}{\textbf{8-shot}} & \textcolor{Purple}{title} & 27.18  & 39.19  & 69.66 \\
  & & \textcolor{Green}{img cap} & 27.25  & 38.61  & 69.67 \\
  & & \textcolor{Purple}{title}+\textcolor{Green}{img cap} & 27.99  & 39.69  & 70.76 \\ 
  & & \textcolor{Purple}{title}+\textcolor{Green}{img cap}+\textcolor{Orange}{OCR text} & 28.80  & 44.10  & 74.71 \\ \hhline{~-----}
  & \textbf{8-shot CoT} & \textcolor{Purple}{title}+\textcolor{Green}{img cap}+\textcolor{Orange}{OCR text}+\textcolor{Blue}{rationale} & 26.32 & 42.06 & 73.95 \\ \hhline{~-----}
  & \multirow{4}{*}{\textbf{12-shot}} & \textcolor{Purple}{title} & 25.71  & 37.15  & 68.26 \\
  & & \textcolor{Green}{img cap} & 25.65  & 36.37  & 68.65 \\
  & & \textcolor{Purple}{title}+\textcolor{Green}{img cap} & 26.63  & 38.57  & 69.96 \\ 
  & & \textcolor{Purple}{title}+\textcolor{Green}{img cap}+\textcolor{Orange}{OCR text} & 28.76  & 43.18  & 73.96 \\ \hhline{~-----}
  & \textbf{12-shot CoT} & \textcolor{Purple}{title}+\textcolor{Green}{img cap}+\textcolor{Orange}{OCR text}+\textcolor{Blue}{rationale} & - & - & - \\ \midrule
  \multirow{8}{*}{\textbf{MiniGPT4}} & \multirow{5}{*}{\textbf{0-shot}} & meme & 06.17  & 22.20  & 63.31 \\
  & & meme+\textcolor{Purple}{title} & 14.37  & 30.70  & 66.19 \\ 
  & & meme+\textcolor{Green}{img cap} & 10.36  & 26.22  & 64.39 \\ 
  & & meme+\textcolor{Purple}{title}+\textcolor{Green}{img cap} & 12.49  & 28.51  & 65.81 \\ 
  & & meme+\textcolor{Purple}{title}+\textcolor{Green}{img cap}+\textcolor{Orange}{OCR text} & 12.46  & 31.44  & 68.62 \\ \hhline{~-----}
  & \textbf{0-shot CoT} & meme+\textcolor{Purple}{title}+\textcolor{Green}{img cap}+\textcolor{Orange}{OCR text}+\textcolor{Blue}{rationale} & 12.57  & 31.70  & 68.45 \\ \hhline{~-----}
  & \multirow{1}{*}{\textbf{finetuned}} & meme+\textcolor{Purple}{title}+\textcolor{Green}{img cap}+\textcolor{Orange}{OCR text} & 7.50  & 27.88  & 65.47 \\ \hhline{~-----}
  & \textbf{FT CoT} & meme+\textcolor{Purple}{title}+\textcolor{Green}{img cap}+\textcolor{Orange}{OCR text}+\textcolor{Blue}{rationale} & 7.25  & 26.68  & 65.86 \\ \bottomrule
    \end{tabular}
    \caption{0, 4, 8, 12 shot results with Flamingo, LLaMA, and MiniGPT4 models. ``-'' indicates the model ran out of memory.}
    \label{tab:full-result}
\end{table*}

\end{document}